\pdfoutput=1

\documentclass[11pt]{article}

\usepackage[final]{coling}

\usepackage{times}
\usepackage{latexsym}

\usepackage[T1]{fontenc}

\usepackage[utf8]{inputenc}

\usepackage{microtype}

\usepackage{inconsolata}

\usepackage{graphicx,textcomp}
\usepackage{booktabs}
\usepackage{multirow}
\usepackage{amsbsy,amsmath}
\usepackage{parskip}
\usepackage{colortbl}
\usepackage[textsize=tiny]{todonotes}
\usepackage{pifont}
\newcommand{\ra}[1]{\renewcommand{\arraystretch}{#1}}
\newcommand{\cmark}{\ding{51}}
\newcommand{\xmark}{\ding{55}}

\title{STTATTS: Unified Speech-To-Text And Text-To-Speech Model}

\author{Hawau Olamide Toyin$^1$, Hao Li$^{1,2}$, Hanan Aldarmaki$^1$ \\ 
        $^1$Mohamed Bin Zayed University of Artificial Intelligence, UAE \\
        $^2$Pinscreen, USA\\
        \texttt{\fontsize{10}{12}\selectfont\{hawau.toyin,hao.li,hanan.aldarmaki\}@mbzuai.ac.ae}\\}

\begin{document}
\maketitle
\begin{abstract}
Speech recognition and speech synthesis models are typically trained separately, each with its own set of learning objectives, training data, and model parameters, resulting in two distinct large networks. We propose a parameter-efficient approach to learning ASR and TTS jointly via a multi-task learning objective and shared parameters. Our evaluation demonstrates that
the performance of our multi-task model is comparable to that of individually trained models while significantly saving
computational and memory costs ($\sim$50\% reduction in the total number of parameters required for the two tasks combined). We experiment with English as a resource-rich language, and Arabic as a relatively low-resource language due to shortage of TTS data. Our models are trained with publicly available data, and both the training code and model checkpoints are openly available for further research.\footnote{\url{https://github.com/mbzuai-nlp/sttatts}}
\end{abstract}

\section{Introduction}
Fundamentally, text and speech are different representations of similar information, with text being a far more condensed form of the linguistic content of speech. Conversion between text and speech modalities in the form Automatic Speech Recognition (ASR),  text-to-speech synthesis (TTS), or Voice Conversion (VC), are traditionally achieved by training separate ASR, TTS, and VC models as the input and output modalities and training objectives differ significantly. However, considering the recent developments in self-supervised and multi-modal pre-training, a more integrated approach can now be promising. A unified model trained simultaneously for multiple speech/text to speech/text, improves on generalization to new data, cross-task knowledge transfer, simplifies maintenance, and also reduces the computational and memory requirements for training, storage, and inference. Recent studies seek to achieve a smooth fusion of text and audio by developing unified audio-text models capable of addressing diverse tasks both within and across these modalities. While these models are considered multi-modal if they can process different input modalities, for our purposes we group audio-text models into two categories based on their output modality: uni-modal and cross-modal. 
We describe uni-modal approaches as models capable of generating output in a single modality only, such as Whisper \cite{whisperradford2022robust}, and Google USM \cite{zhang2023googleusm}, which only generate text outputs. Cross-modal approaches, on the other hand, are capable of generating outputs in both speech and text modalities, such as Viola \cite{wang2023viola}, SpeechT5 \cite{ao2022speecht5}, and SpeechGPT \cite{zhang2023speechgpt}.
Some of these models \cite{zhang2023speechgpt, maiti2024VoxtLM} use discrete representations for audio tasks, merging text and audio tokens in a shared vocabulary while jointly training multiple tasks; other models, such as SpeechT5 \cite{ao2022speecht5} use continuous representations for audio. While SpeechT5 is pre-trained with a cross-modal objective, handling both text and speech as input and output modalities, the model is fine-tuned individually for downstream tasks such as ASR and TTS. 

This work builds on these developments, particularly SpeechT5 \cite{ao2022speecht5}, by achieving a truly cross-modal speech and text conversion in a single architecture. Unlike previous work, our approach does not separate the speech/text-to-speech/text tasks based on input/output modalities; instead, we fine-tune these components concurrently using a unified model and loss function with the help of a simple MLP-based task fusion module. The resulting model is a single encoder-decoder that can handle both modalities at the input and output, depending on the desired task. We summarize our contributions below: 
\begin{enumerate}
    \item We propose a novel parameter-efficient fine-tuning methodology for jointly learning multiple speech tasks: \textbf{ASR and Multi-Speaker TTS}. We demonstrate the  \textbf{efficiency} of our approach in terms of computational requirements, training time, and its \textbf{scalability} to additional tasks, namely \textbf{Voice Conversion}.
    \item We empirically demonstrate the effectiveness and efficiency of the proposed approach, resulting in \textbf{improved performance} compared to the only comparable open-source model at the time of writing (i.e. VoxLM) with a fraction (1/2) of the parameters. 
    \item We demonstrate the \textbf{robustness} and \textbf{performance} of our approach, showing its applicability on both high-resource and \textbf{low-resource} settings, as we present the \textbf{first multi-modal, multi-task model} for the \textbf{Arabic} language.
    
\end{enumerate}

\section{Related Work}
\subsection{Multi-Task Speech Models} 

\citeauthor{whisperradford2022robust} introduced Whisper, an encoder-decoder model trained on vast amount of speech-text (680K hours) data. Whisper a multilingual model with multitasking capabilities. However, it only uses speech as input and can not generate speech output.  In SLAM \cite{bapna2021slam}, the authors unified speech and text pre-training within a single model using a single encoder with the combined BERT \cite{devlin2019bert} and W2V-BERT \cite{chung2021w2vbert} objectives on unlabeled text and speech. To align their model’s representations across modalities, they used Translation Language Modeling (TLM) and Speech Text Matching (STM) alignment losses that use supervised speech-text recognition data. They show that joint pre-training improves model performance on downstream speech translation and recognition tasks. However, these models are not jointly trained for multi-task purposes and require dedicated fine tuning for each task.

\subsection*{Unified Speech-Text Models} 

SpeechT5 \cite{ao2022speecht5} introduces a multi-modal encoder-decoder pre-training approach for spoken language processing. The authors attempted a joint pre-training approach of speech and text to improve the model's performance on downstream speech/text tasks like ASR, TTS, speaker identification, speech enhancement, and voice conversion. They built on the transformer architecture \cite{NIPS2017_attention}, adding modal-specific pre-nets and post-nets to handle latent feature extraction/conversion for different modalities. Although their model is pre-trained jointly with speech and text data in a self-supervised manner, the supervised downstream models (e.g. ASR, TTS) were \emph{trained individually for each task}. The model was trained and evaluated on English only. A subsequent work followed the same architecture and training paradigm for building an Arabic version of the model, named ArTST \cite{toyin-etal-2023-artst}, which also requires task-specific training.

Some recent methods employ a decoder-only framework post-conversion of continuous audio into discrete tokens, subsequently combining text and audio tokens into a unified vocabulary \cite{maiti2024VoxtLM, zhang2023speechgpt, wang2023viola}. These models can generate both text and speech output from speech/text input. Some models \cite{maiti2024VoxtLM} discretize speech using k-means on features extracted from speech-text pre-trained models like HuBert \cite{hsu2021hubert}. However, their method can suffer from information loss caused by quantizing speech signals into discrete tokens and its performance highly depends on the value of \emph{k} used for feature extraction. Moreover, combining text and discrete speech tokens into a vocabulary can lead to a large vocabulary size for multilingual training.

\section{Our Method}
In this section, we describe a unified  \textbf{S}peech-\textbf{T}o-\textbf{T}ext \textbf{A}nd \textbf{T}ext-\textbf{T}o-\textbf{S}peech model, \textbf{STTATTS},  our proposed architecture for jointly training ASR and TTS.  After unified self-supervised training as described in \citet{ao2022speecht5}, we propose utilizing a multi-task loss objective to optimize our model parameters for multiple tasks, along with a task fusion module to handle the different tasks. 
Unlike the fine-tuning methodology followed in SpeechT5 \cite{ao2022speecht5, toyin-etal-2023-artst}, which results in a completely disjoint copy of the model for each task (see Figure 2: \textit{Right}), \texttt{STTATTS} utilizes a task fusion module with negligible number of additional parameters to handle multiple tasks using the same encoder-decoder backbone (See Figure 1). This results in a $\sim$50\% reduction in the total number of parameters required for the two tasks combined.

\begin{figure}[t]
  \centering
  \includegraphics[width=1.0\columnwidth]{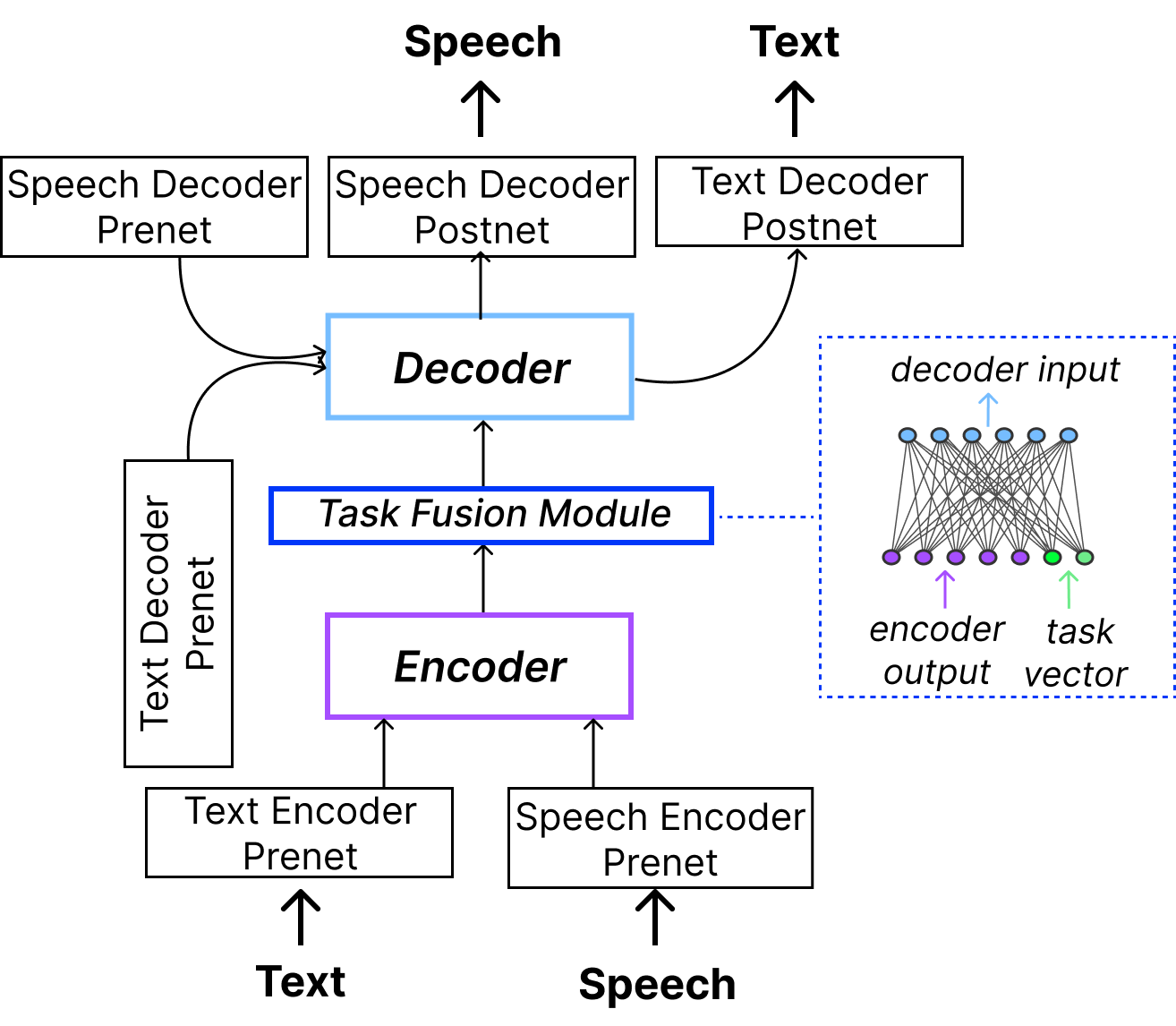}
  \caption{\texttt{STTATTS} architecture: the task fusion module is used to condition the encoder output to a specific task.}
  \label{fig:our_model}
\end{figure}

\begin{figure*}
    \centering
    \includegraphics[width=0.8\linewidth]{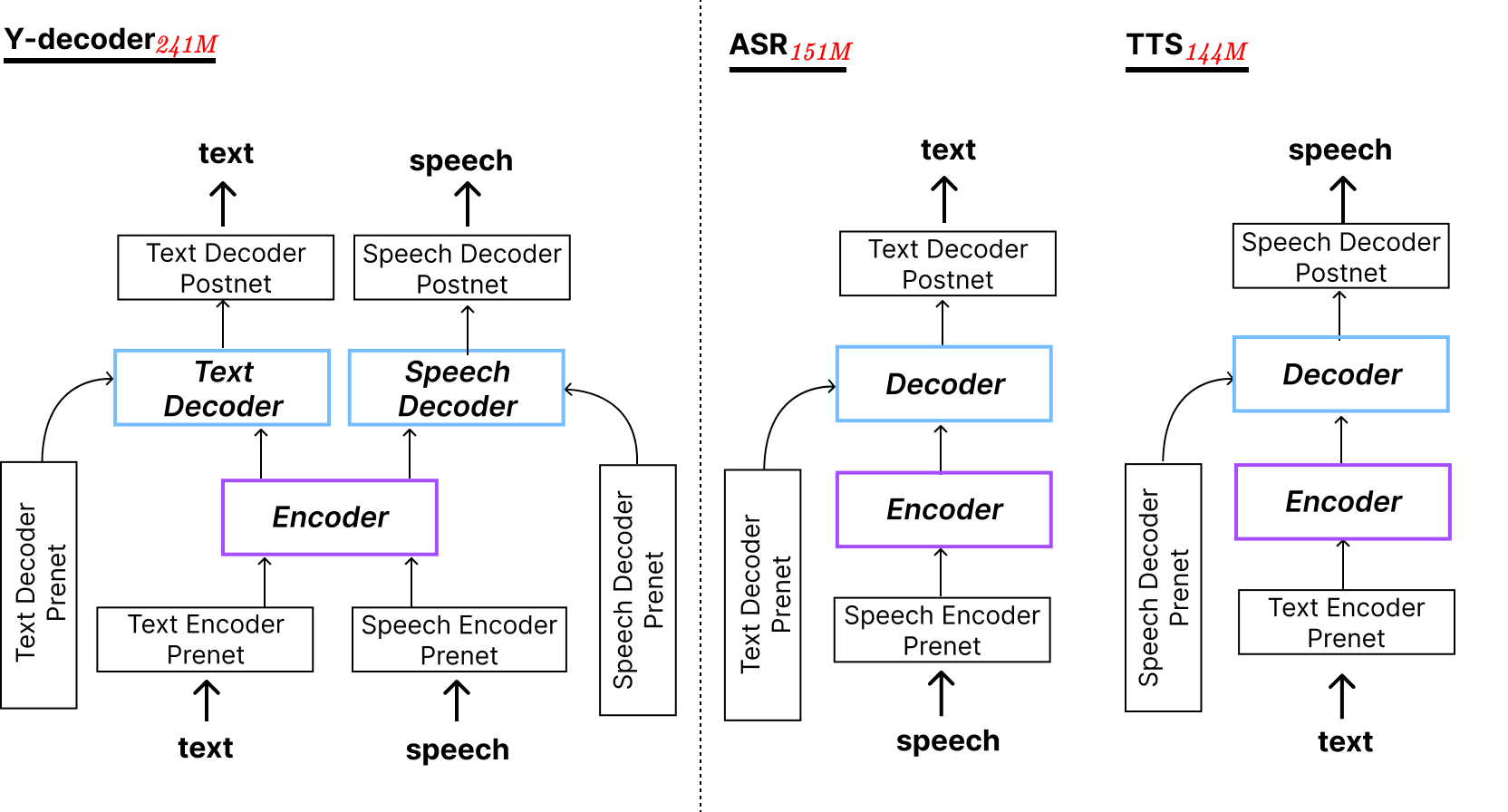}
    \caption{\emph{Left:} Other modifications we experimented with. \emph{Right:} Architecture for single task training using the SpeechT5/ArTST methodology.}
    \label{fig:ablation}
\end{figure*}
\subsection{Underlying Architecture}
The model is based on the SpeechT5 architecture, which comprises transformer encoder-decoder blocks and some auxiliary modules for modality-specific feature extraction and decoding. Specifically, the modal-specific encoder prenets, and modal-specific decoder pre/postnet modules, are used to handle the text and speech modalities at the input and output, while the main encoder/decoder network processes the unified representations.

\paragraph{Unified Encoder-Decoder.}

The unified encoder-decoder follows the transformer architecture \cite{NIPS2017_attention}. The transformer encoder in our \texttt{base} model has 12 blocks, with a model dimension of 768 and an inner feed-forward network dimension of 3072. The decoder comprises 6 transformer decoder blocks with a model dimension of 768 and an inner dimension of 3072. The unified encoder and decoder can process either latent representation of text or speech. We use the pre-trained encoder weights from SpeechT5 and ArTST for English and Arabic experiments, respectively. 

\paragraph{Text Encoder and Decoder Pre/Post-Nets.}
The text encoder pre-net and text decoder pre/post-nets use shared embeddings. The
pre-nets transform a token in the sequence to a 768 embedding vector. The text decoder post-net projects the decoder’s hidden state into the probability distribution of tokens, normalized by the softmax function.

\paragraph{Speech Encoder and Decoder Pre/Post-Nets.}

The speech encoder pre-net is a convolutional feature extraction model
with 6 1-dimensional convolutional layers, and GELU activation following wav2vec2.0 \cite{wav2vec}. The output of the convolutional network is normalized before passing to a linear layer to up-sample from 512 to 768. 
The speech decoder pre-net upsamples the mel spectrogram to a 768-dimensional vector. It consists of two sequential layers of Linear transformations with ReLU activations, followed by an additional Linear layer that upsamples the output from the previous layers.  To enable multi-speaker TTS, the speaker embedding vector of the target speaker extracted using x-vectors \cite{x-vectors} is concatenated to the output of the speech decoder pre-net before being down-sampled with a linear layer to the decoder hidden size followed by RELU activation. 

The speech decoder post-net utilizes a linear layer to predict the log mel-filterbank from the decoder output and five 1-dimensional convolutional layers to generate a residual for enhancing the predicted mel \cite{ao2022speecht5}. Another linear module is incorporated to transform the decoder output into a scalar to predict the stop token.
\citeposs{ao2022speecht5} HiFi-GAN vocoder is used to synthesize speech from the generated mel spectrogram.

\subsection{Task Fusion Module}  
Each task is represented with a 128-dimensional vector, which is concatenated with the encoder's output, followed by a fully connected layer (See Figure \ref{fig:our_model}) to project the learned representation back to the encoder embedding size. The output of this module is used as input to the decoder.

\subsection{Multi-Task Loss Objective}

For ASR, we use $\mathcal{L}_{ce}$: the cross-entropy loss of the decoder and $\mathcal{L}_{ctc}$: the standard CTC loss from ESPNet \cite{espnet}.

\begin{equation}
    \mathcal{L}_{asr} = \mathcal{L}_{ce} + \mathcal{L}_{ctc}
\end{equation}\label{eq:tts}

For optimizing speech synthesis, we use the $\mathcal{L}_1$ loss to minimize the distance between the target and generated mel spectrograms; the binary cross entropy $\mathcal{L}^s_{bce}$ loss is used to predict the stop token for generation; and the guided attention loss is used to speed up training convergence for speech synthesis as described by \citet{8461829}. The latter was added to speed up the training of TTS \cite{ao2022speecht5} which is typically a lot slower compared to ASR. 
\begin{equation}
    \mathcal{L}_{tts} = \mathcal{L}_{1} + \mathcal{L}_{bce} + \mathcal{L}_{attn}
\end{equation}

For joint training, similar convergence rates enable better and consistent training.
The ASR and TTS objectives are combined as \( \mathcal{L} = \mathcal{L}_{asr} + \mathcal{L}_{tts}\). At each step, we calculate the loss for each task and normalize by the number of samples per task in that step. The model is updated after some $k$ gradient accumulation steps.

\paragraph{Input/Output Representation.}
For speech, we use the raw waveform sampled at 16kHz as input and the
80-dimensional log mel-filterbank features as target output for all experiments. We trained with a maximum input token size of 480K, which corresponds to 30 seconds of speech.
For text, character-level tokens served as input and output for all experiments. The maximum input token length is 600. Our vocabulary size is 98 for experiments using Arabic following ArTST \cite{toyin-etal-2023-artst} and 84 for English following SpeechT5 \cite{ao2022speecht5}.

\section{Experimental Settings}

We conducted experiments across two languages: English and Arabic using open-source datasets. English was used as a resource-rich language, and trained using the benchmark LibriSpeech dataset; this  enables the utilization of pre-trained models from SpeechT5 \cite{ao2022speecht5}, and a direct comparison with their downstream models. Arabic served as a relatively low-resource language, mainly because of the shortage of clean speech data that can be utilized for training high-quality TTS systems \cite{kulkarni2023clartts}. This also enables using the pre-trained checkpoints from ArTST \cite{toyin-etal-2023-artst}.

\subsection{Datasets}
\paragraph{English.} 
\noindent We used LibriSpeech \cite{Librispeech}, referred to as \verb|LS| for ASR. For TTS, we use LibriTTS \cite{zen2019libritts}, referred to as \verb|Ltts|, along with LJSpeech \cite{ljspeech17}. We conducted experiments with varying training data sizes to evaluate the impact of data size on performance. See Table \ref{tab:english-datasets} for data combinations.

\begin{table}[ht]
\centering
\ra{1.3}
\resizebox{\columnwidth}{!}{%
\begin{tabular}{@{}l*{7}{c}@{}}
\toprule
 & \multicolumn{3}{c}{\textbf{LS}} & \multicolumn{3}{c}{\textbf{Ltts}}  & \textbf{LJSpeech} \\ \cmidrule(lr){2-4} \cmidrule(lr){5-7} 
  & 100hr & 360hr & 500hr & 100hr &  360hr & 500hr & - \\ \midrule
$en_s$ & \cmark & \xmark & \xmark & \cmark  & \cmark  & \xmark & \xmark \\
$en_m$ & \cmark & \cmark & \xmark  & \cmark & \cmark & \cmark &\xmark \\
$en_l$ & \cmark & \cmark & \cmark  & \cmark& \cmark & \cmark & \cmark \\  
 \bottomrule
\end{tabular}%
}
\caption{Datasets for English experiments.}
\label{tab:english-datasets}
\end{table}

\paragraph{Arabic.} \label{arabic_data}
We combined quality data for speech synthesis from two publicly available datasets: Arabic Speech Corpus (ASC) \cite{halabi2016modern}, and Classical Arabic Text-to-Speech Corpus (ClArTTS) \cite{kulkarni2023clartts} for both ASR and TTS.
Since TTS quality is highly sensitive to the quality and consistency of the audios and annotations used for training, we did not utilize other large speech data for joint training \footnote{Our preliminary experiments showed that balancing training data for TTS and ASR are needed to achieve balanced outcomes on both tasks. As TTS data are limited for Arabic, we limited the data used for ASR training to achieve that balance.}. This also serves as a good setting for validating the approach on low-resource settings.  We combined the train and test sets of both datasets. However, the original transcripts for the ASC corpus were modified to match the phones \cite{halabi2016modern}, leading to the removal/addition of characters that are silenced/pronounced. These modifications negatively impact both ASR and TTS performance in our set-up as they are inconsistent with the transcripts used in ClArTTS, which follows standard Arabic spelling conventions. To rectify this issue, we restored the transcripts to standard Arabic spelling. We utilized ChatGPT 4.0 to restore the spelling, and manually inspected a subset of the resulting transcripts, which were deemed to be correct by a native Arabic evaluator. We used regex to remove punctuation marks, newline characters, and additional English text from the model's output. The prompt used was: \emph{ In the given arabic texts, fix the incorrect characters in the words, take each line as a sentence, and return the same number of lines as passed. Don't remove the diacritics \{line separated transcriptions\} }.

\subsection{Data Preparation}
All punctuation marks were removed for both English and Arabic texts and all English characters were converted to lowercase letters. The standardized sampling rate for speech data across all collected datasets was 16 kHz. For Arabic, we trained the model for TTS with and without diacritics.

\subsection{Training Details}
The pre-trained checkpoints from SpeechT5\footnote{\url{https://github.com/microsoft/SpeechT5/tree/main/SpeechT5}} and ArTST\footnote{\url{https://github.com/mbzuai-nlp/ArTST}} were used as initialization weights for our experiments on English and Arabic, respectively. For reproducibility, we summarize training details per experiment in Table \ref{tab:experiment_details}. All experiments were carried out on A100 GPUs with 80GB memory. Our training code and checkpoints are available.\footnote{\url{https://github.com/mbzuai-nlp/sttatts}}

\paragraph{Warm Fine-tuning.} \citet{toyin-etal-2023-artst} discuss that in the low-resource TTS setting for Arabic, fine-tuning with larger ASR datasets first, followed by continual fine-tuning with high-quality TTS data improved synthesized speech quality. The resulting model achieved higher intelligibility even without the use of diacritics. We incorporate this finding in our Arabic model by first fine-tuning the model for single-task TTS on the MGB2 \cite{ali2016mgb} dataset, followed by continual fine-tuning for joint ASR and TTS using our smaller dataset, which improves TTS performance.
For English, we utilize a similar approach for a different purpose: In $en_l$, the sizes of ASR and TTS training sets are imbalanced, which negatively impacts TTS performance. In the first 200k updates, we use the train-other-500 split from \texttt{LS} \cite{Librispeech} for optimizing the ASR part of the joint objective. For the rest of the training duration, we use the rest of the data (the combined \texttt{LS} 100 and \texttt{LS} 360) to continue optimizing ASR in the joint objective. The training data for TTS (see Table \ref{tab:english-datasets}) remain unchanged throughout training. This approach enables the utilization of the full ASR train set without negatively impacting TTS performance. 
Ablation results with and without this approach are discussed in Section \ref{sec:warm-finetune}. 
\begin{table}[h]
\centering
\scalebox{0.75}{%
\begin{tabular}{@{}lcc@{}}
\toprule
\multicolumn{1}{c}{} &
  \multicolumn{1}{c}{Arabic} &
  \multicolumn{1}{c}{English ($\boldsymbol{\mathcal{D}^{en}_{s}}$)} \\ \hline
\begin{tabular}[c]{@{}l@{}}\emph{Data}\\ \hspace{0.5em}Total hours\\ \hspace{3mm} - ASR\\ \hspace{3mm} - TTS\\\hspace{0.5em} Character vocabulary size\end{tabular} &
  \begin{tabular}[c]{@{}c@{}}-\\ 32\\ 16\\ 16\\ 98\end{tabular} &
  \begin{tabular}[c]{@{}c@{}}-\\ 291 \\ 100\\ 191\\ 84\end{tabular} \\ \midrule
\begin{tabular}[c]{@{}l@{}}\emph{Model architecture}\\\hspace{0.5em} Parameters (M)\end{tabular} &
  \begin{tabular}[c]{@{}c@{}}-\\ 155\end{tabular} &
  \begin{tabular}[c]{@{}c@{}}-\\ 155\end{tabular} \\ \midrule
\begin{tabular}[c]{@{}l@{}} \emph{Training Configurations}\\\hspace{0.5em}  Max. input tokens (M)\\\hspace{0.5em}  Total updates (K)\\ \hspace{0.5em} Update frequency ($k$) \\ \hspace{0.5em}  Learning rate\\ \hspace{0.5em} LR scheduler\\ \hspace{0.5em} phase ratio\\ \hspace{0.5em}  Optimizer \\ \hspace{0.5em}  \# GPU \end{tabular} &
  \begin{tabular}[c]{@{}c@{}}-\\ 4.0 \\ 80 \\ 6 \\ 1e-4\\ tri-stage\\ 0.25,0.4,0.35\\ adam \\ 3 \end{tabular} &
  \begin{tabular}[c]{@{}c@{}}-\\ 3.2\\ 150\\ 8 \\ 1e-4\\ tri-stage\\ 0.25,0.4,0.35\\ adam \\ 3 \end{tabular} \\ \bottomrule
\end{tabular}%
}
\caption{Experiment details. For $en_m$ we use 250K training updates and 4 A100 GPU's. For $en_l$ we use 450K training updates and 4 A100 GPU's.}
\label{tab:experiment_details}
\end{table}

\subsection{Evaluation Metrics}
\paragraph{ASR.}
We report the Word and Character Error rates (WER \& CER) for evaluating our model's performance for ASR task. For Arabic, the texts are normalized before evaluation by removing diacritics. Diacritics are mostly useful as input for TTS models, but most ASR models are trained and evaluated without diacritics. We transcribe speech using CTC weight of 0.5.

\paragraph{TTS.}
We employ objective metrics to evaluate our model's synthesized speech intelligibility. We use Whisper's \cite{whisperradford2022robust} large model to transcribe synthesized speech and calculate CER of transcribed speech. Subjective Mean Opinion Score (MOS) was used to measure naturalness and intelligibility for the main results, in addition to predicted MOS values using Wav2Vec2.0 as described in \citet{Andreev_wvmos_2023}. Native speakers of both languages rated 5 random samples of synthesized speech on a scale of 1 to 5, with a step of 0.5. 

\section{Results}
In this section, we report our model's performance, comparison against SpeechT5 single-task models (see Figure \ref{fig:ablation}) and comparison against a baseline joint-task model, VoxLM. 
Table \ref{tab:our-results} shows the results for Arabic and English with various scales of training data. The performance improves with additional training data, approaching state-of-the-art results on the LibriSpeech test set.  
On Arabic, the high WER is attributed to the low-resource setting as it was trained with only 16 hours for ASR. 
Examples of ASR predictions from the Arabic model are shown in Figure \ref{fig:arabcic-asr-trasncript}.

\begin{table}[h]
\centering
\ra{1.2}
\resizebox{1\columnwidth}{!}{%
\begin{tabular}{@{}l cc cccc@{}}
\toprule
\multirow{2}{*}{\textbf{Data}} & \multicolumn{2}{c}{\textbf{ASR}} & \multicolumn{4}{c}{\textbf{TTS}}  \\ \cmidrule(lr){2-3} \cmidrule(l){4-7} 
 & WER $\downarrow$ & CER $\downarrow$ & CER $\downarrow$& Naturalness $\uparrow$ & Intelligibility $\uparrow$ & WV-MOS $\uparrow$ \\ \midrule
$ar$ & 10.22 & 2.63  & 6.22  & 3.28 & 2.78 & 3.69 \\
$en_s$& 4.84 & 1.63 & 3.18 & 3.36 & 4.00 & \textbf{4.40} \\
$en_m$& 3.47 & 1.07 & 2.74 & 3.20 & 4.04 & 4.24\\
$en_l$& \textbf{2.99} & \textbf{0.90}& \textbf{2.10} & 3.00 & \textbf{4.38} & 4.26\\ \bottomrule
\end{tabular}%
}
\caption{ Evaluation of ASR performance using WER and CER on LibriSpeech test set, and TTS performance using CER, Naturalness MOS, Intelligibility MOS, and MOS scores predicted by Wav2Vec2.0 (WV-MOS).
}
\label{tab:our-results}
\end{table}

\begin{table}[h]
\centering
\ra{1.3}
\resizebox{1\columnwidth}{!}{%
\begin{tabular}{@{}lccc@{}}
\toprule
\large
\hspace{0.5em}\multirow{2}{*}{\textbf{Model}} &
  \multirow{2}{*}{\textbf{\# params}(M)} &
  \textbf{ASR} &
  \textbf{TTS} \\ 
  \cmidrule(lr){3-3}\cmidrule(l){4-4}
 & &
 WER $\downarrow$ &
  CER $\downarrow$ \\ \midrule
\textbf{\emph{Arabic}} &&& \\
\hspace{0.5em}$^\star$ArTST$_{ASR}$ \cite{toyin-etal-2023-artst} & 151 & \textbf{7.59} & $\times$ \\ 
\hspace{0.5em}$^\star$ArTST$_{TTS}$ & 145 & $\times$ & 9.61 \\ 
\hspace{0.5em}$^\star$Whisper$_{small}$ \cite{whisperradford2022robust} & 244 & 32.2 & $\times$ \\ 
\hspace{0.5em}$^\star$Whisper$_{large}$ & 1550 & 23.4 & $\times$ \\ 
\hspace{0.5em}\texttt{STTATTS} & 155 & 10.22 & \textbf{6.22} \\ \midrule
\textbf{\emph{English $en_s$}} &&& \\
\hspace{0.5em}$^\dagger$SpeechT5$_{ASR}$ \cite{ao2022speecht5} & 151 & 4.4 & $\times$ \\ 
\hspace{0.5em}$^{\dagger\dagger}$SpeechT5$_{TTS}$ & 145 & $\times$ & 6.3 \\ 
\hspace{0.5em}$^\dagger$Whisper$_{small}$ & 244 & 3.4 & $\times$ \\ 
\hspace{0.5em}$^\dagger$Whisper$_{large}$ & 1550 & \textbf{3.0} & $\times$ \\ 
\hspace{0.5em}\texttt{STTATTS} & 155 & 4.8 & \textbf{3.2} \\ \bottomrule
\end{tabular}
}
\caption{Comparison of \texttt{STTATTS} with single task models on our test sets. Whisper models use 438K hours of English ASR data, and SpeechT5 uses 100 hours LibriSpeech for ASR and 460 hours LibriTTS for TTS. For ArTST results, we train the model with our combined dataset and report evaluation results on our test set. $^\star$ indicates we trained the models and evaluated ourselves, $^\dagger$ indicates results as reported in the corresponding paper, $^{\dagger\dagger}$ indicates the evaluation of model on randomly selected synthesized text as ours, \textbf{$\times$} indicates the models cannot perform the corresponding task.}
\label{tab:STTATTSvssingletask}
\end{table}

\subsection{Comparison with Single-Task Models}
\noindent In Table \ref{tab:STTATTSvssingletask}, we compare our model with single task models from ArTST \cite{toyin-etal-2023-artst} and SpeechT5 \cite{ao2022speecht5}, which have the same pre-trained checkpoint and fine-tuning data, so they are directly comparable: we used the \textbf{$en_s$} described in Table \ref{tab:english-datasets} for English and \textbf{$ar$}. \texttt{STTATTS} achieves comparable performance to the single-task models, demonstrating the effectiveness of the proposed multi-task methodology. Compared to the large-scale multi-lingual model, Whisper, \texttt{STTATTS} performs better than Whisper$_{small}$, in spite of having smaller number of parameters and being trained on a much smaller data set. In English,  Whisper$_{large}$ performs marginally better than \texttt{STTATTS} trained on  \textbf{$en_s$}. For Arabic, \texttt{STTATTS} performs substantially better than both small and large variants.

\subsection{Comparison with Joint-Task Models}
We compare our model's performance with reported joint models that perform both TTS and ASR in Table \ref{tab:STTATTS_vs_joint_task}. In particular, we use VoxtLM \cite{maiti2024VoxtLM} as it is publicly available for direct comparison, while other models are not currently available to perform comparable evaluation. 

\noindent Using the same training data, \texttt{STTATTS} performs better in both tasks, even when compared with the large VoxtLM model of 1.3B parameters. Compared to the VoxtLM base, which has a comparable number of parameters, \texttt{STTATTS} performs on a par with a fraction of the training data and is far better when an almost equal amount of data is used. 

\begin{table}[h]
\centering
\ra{1.2}
\resizebox{1.0\columnwidth}{!}{%
\begin{tabular}{@{}lcccc@{}}
\toprule
\multirow{2}{*}{\textbf{Model}} & \multirow{2}{*}{\textbf{\# params}} & \textbf{Train Data(hrs)} & \textbf{ASR} & \textbf{TTS} \\ \cmidrule(lr){3-3}  \cmidrule(lr){4-4}  \cmidrule(l){5-5}
 & & \multicolumn{1}{c}{ASR/TTS} & \multicolumn{1}{c}{WER$\downarrow$} & CER$\downarrow$ \\ \midrule

\textbf{\emph{English}} \\
\hspace{0.5em} VoxtLM\_base$\dagger$ \citeyear{maiti2024VoxtLM} & 350M & 0.9K/0.5K & 6.5 & 3.5 \\
\hspace{0.5em} VoxtLM\_large$\dagger$ & 1.3B & 0.9K/0.5K & 4.6 & 3.9 \\
\hspace{0.5em} VoxtLM\_large$\dagger$ & 1.3B & 45K/0.6K & \underline{2.7} & 3.6 \\
\hspace{0.5em} \texttt{STTATTS($en_s$)} & 154M & 0.1K/0.2K & 4.8 & 3.2 \\
\hspace{0.5em} \texttt{STTATTS($en_l$)} & 154M & 0.9K/0.5K & \textbf{3.0} & \textbf{2.1} \\ \bottomrule
\end{tabular}%
}
\caption{Comparison with other joint-task models. $\dagger$ indicates results as reported in the reference paper. ASR results are on the Librispeech test-clean set for all models. TTS results are on 100 subset from the LibriTTS test set, we can't compare with the same samples since the samples used for testing in VoxtLM are not publicly available.
}
\label{tab:STTATTS_vs_joint_task}
\end{table}

\section{Ablations \& Analysis} 

\subsection{Task Scaling}
To evaluate whether the model can be scaled to perform additional speech/text to speech/text tasks, we experiment with adding Voice Conversion in the $en_m$ setting. We use the CMU Arctic \cite{kominek04b_cmuarctic} dataset, and optimize voice conversion using the same loss function used for TTS (see Equation \ref{eq:tts}). We evaluate VC performance using CER and Speaker Similarity (SS) using the cosine similarity function on speaker embedding \cite{x-vectors} extracted from the speech.
The results are shown in Table \ref{tab:scale_tasks_results}. Although there's a slight reduction in WER for the ASR task, TTS MOS score has improved, possibly due to the shared output space and loss function. We achieve speaker similairy score of 0.99, and a low CER, demonstrating the high quality in the additional VC task without any additional parameters added to the model.  

\begin{table}[h]
    \centering
    \ra{1.2}
    \resizebox{\columnwidth}{!}{
    \begin{tabular}{@{}lcccccccc@{}} \toprule
          & \multicolumn{2}{c}{\textbf{ASR}} & \phantom{a} & \multicolumn{2}{c}{\textbf{TTS}} &  \phantom{a} &\multicolumn{2}{c}{\textbf{VC}} \\ \cmidrule{2-3} \cmidrule{5-6} \cmidrule{8-9}
       \textbf{Data}  & WER & CER & & CER & WV-MOS & & CER & SS \\ \midrule
        $en_m$ & 3.59 & 1.13 && 2.83 & 4.28 && 1.58 & 0.99 \\
        \bottomrule
    \end{tabular}
    }
    \caption{Results for joint training of ASR,TTS and VC.}
    \label{tab:scale_tasks_results}
\end{table}

\subsection{Architectural Variations}

\paragraph{Y-Decoder.} \textbf{Y} in \textbf{Y}-decoder stands for the desired output: either text or speech. Here, we use a similar approach to \texttt{STTATTS} but with modal-specific decoders, i.e., a separate text decoder and a speech decoder, while sharing the same encoder. In this model, no task fusion module is used since each task is parameterized by its own standalone decoder. This results in a larger model than \texttt{STTATTS} but still smaller than the disjoint ASR and TTS SpeechT5 models.  See Figure \ref{fig:ablation} (left) for an illustration of Y-decoder architecture. 

\paragraph{Multi-Stage Training.} We also experimented with multi-stage training, where we alternate updating the weights for a specific task with some model components frozen/unfrozen. We performed two experiments using \texttt{STTATTS}, where we update the parameters of the ASR first, followed by TTS, or vice versa, with the respective single-task loss function in each stage. For example, if we follow an ASR-first schedule, we first update the auxiliary weights, the encoder, and the decoder for the ASR task, then in the second stage, we keep the encoder and decoder frozen and update the auxiliary weights for the TTS task.

\paragraph{Task-Specific Adapters.} We experimented with a shared transformer encoder-decoder backbone using pre-trained weights with adapters \cite{pmlr-v97-houlsby19adapter} as an alternative architecture. We tried using the base model architecture offline and only updating the weights of an adapter for each task. We added a 64 inner dimension adapter to the decoder, but this approach didn't yield good results. For ASR, the WER was 150\% and the TTS loss converged at a high value ($\approx0.8$) relative to \texttt{STTATTS} ($\approx0.4$).

\paragraph{Results.} 
Evaluation results are shown in Table \ref{tab:ablation_results}. While performance on ASR is comparable across models and slightly better with the Y-decoder architecture, \texttt{STTATTS} is the best overall in terms of balancing performance across multiple tasks while maintaining a lower number of parameters. With multi-stage training, we find that it works better when we update the encoder weights for the ASR task first. We observe TTS performance is good whether the transformer encoder is optimized for ASR or TTS first, but the performance for ASR significantly degrades with $\times10$ difference between the WER for ASR-first and TTS-first approaches.  This may be attributed to the dimensionality of the input features for each task, as ASR requires more computations to process the input in the encoder, while TTS works with discrete text input.

\begin{table}[h]
    \centering
    \ra{1.3}
    \resizebox{\columnwidth}{!}{
    \begin{tabular}{@{}lccccc@{}} \toprule
  \multirow{2}{*}{\textbf{Model}}   & \multirow{2}{*} {\textbf{\# params}(M)} &  \multicolumn{2}{c}{\textbf{ASR}}  & \multicolumn{2}{c}{\textbf{TTS}}  \\ \cmidrule(lr){3-4} \cmidrule(lr){5-6}
         &  & WER$\downarrow$ & CER$\downarrow$ & CER$\downarrow$ & WV-MOS$\uparrow $ \\ \midrule
    \textbf{\emph{Arabic}} &  &  &  &  & \\
    \hspace{0.5em} Multistage  &  &  &  &  & \\
    \hspace{1em} TTS-first   & 155 & 109.94 & 78.47 & 9.61 & 3.70 \\
    \hspace{1em} ASR-first  & 155 & 11.60 & 7.80 & 67.87 & 2.79 \\
    \hspace{0.5em} Y-decoder (enc - 2$\times$ dec) & 211 & 10.37 & 2.78 & 8.31 & 3.68 \\
    \hspace{0.5em} \texttt{STTATTS} & 155 & 10.22 & \textbf{2.63} & \textbf{6.22} & \textbf{3.69} \\ \midrule
    \textbf{\emph{English $en_s$}} &  &  &  &  & \\ 
    \hspace{0.5em} Y-decoder (enc - 2$\times$ dec) & 211 & 5.67 & 1.91 & 4.36 & 4.45 \\
    \hspace{0.5em} \texttt{STTATTS} & 155 & \textbf{4.84} & \textbf{1.63} & \textbf{3.18} & \textbf{4.40} \\ 
    \bottomrule
    \end{tabular}
    }
\caption{Results from other architectural variation experiments. For \emph{Multistage}, we only show our preliminary results for Arabic since the performance was sub-optimal.}
\label{tab:ablation_results}
\end{table}

\begin{figure*}[t]
    \centering
    \includegraphics[width=1.0\linewidth]{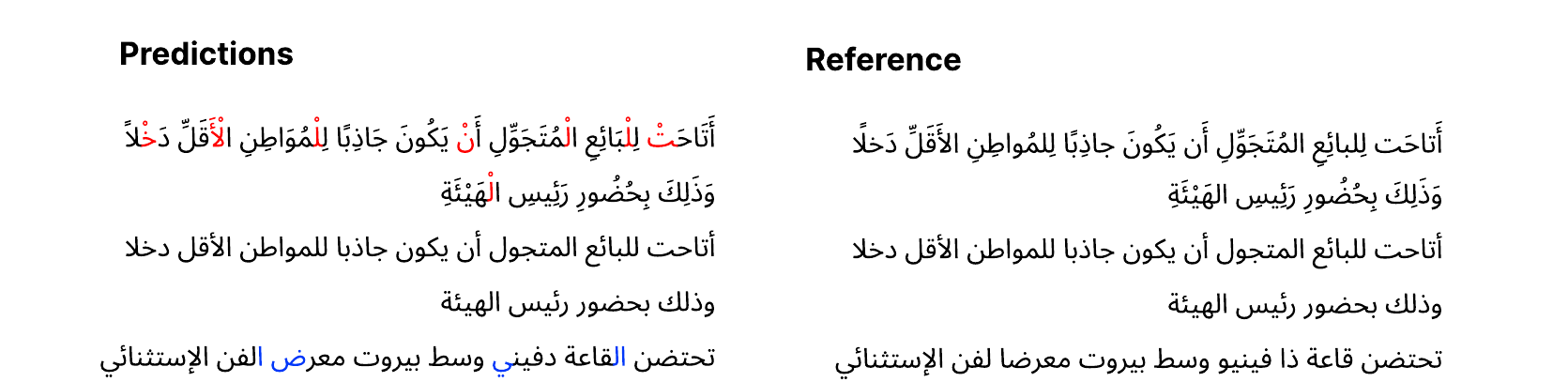}
    \caption{Sample ASR predictions from the Arabic \texttt{STTATTS} model. \textcolor{red}{Diacritic Errors} \textcolor{blue}{Character Errors}}
    \label{fig:arabcic-asr-trasncript}
\end{figure*}

\subsection{Effect of Data Imbalance}
For $en_{s}$, we first fine-tuned with \texttt{LS}-100 and \texttt{Ltts}-100 (which contains $\approx58$ hours of speech). This data combination resulted in less intelligible and robotic synthesized speech (MOS of 1.5). Interestingly, the ASR performance is not affected when we have more TTS data, as shown for $en_s$ in Table \ref{tab:our-results}; on the contrary, the results improved from 5.61 to 4.84. As a result, and since TTS data are generally more scarce, our main experimental settings are all conducted with ASR data downsampled to match the size of the TTS training set. 

\subsection{Effect of Warm Fine-Tuning}
\label{sec:warm-finetune}
We compare \texttt{STTATTS}'s performance with and without the warm fine-tuning approach introduced described in section \ref{sec:warm-finetune}. The results are shown in Table \ref{tab:warm_restart_ar}. Except for the Arabic ASR performance that is degraded by $\approx$ 2\% absolute WER, we find that this approach results in improved performance. 

\begin{table}[h]
\centering
\ra{1.2}
\resizebox{0.9\columnwidth}{!}{%
\begin{tabular}{@{}l c cc cc@{}}
\toprule
\multirow{2}{*}{\textbf{Data}} & \multirow{2}{*}{\textbf{Warm ft}} & \multicolumn{2}{c}{\textbf{ASR}} & \multicolumn{2}{c}{\textbf{TTS}}  \\ \cmidrule(lr){3-4} \cmidrule(l){5-6} 
 & & WER $\downarrow$ & CER $\downarrow$ & CER $\downarrow$ & WV-MOS $\uparrow$\\ \midrule
$ar$ & \xmark & 8.61  & 5.60 & 9.94 & 3.61  \\ 
$ar$ & \cmark & 10.22 & \textbf{2.63}  & \textbf{6.22} & \textbf{3.69}   \\
$en_l$ & \xmark & 3.08 & 0.96 & 3.28 & 4.24 \\
$en_l$ & \cmark & \textbf{2.99} & \textbf{0.90}& \textbf{2.10} & \textbf{4.26} \\
 \bottomrule
\end{tabular}%
}
\caption{Results of experimenting with warm fine-tuning (ft) and without.}
\label{tab:warm_restart_ar}
\end{table}

\subsection{Diacritization in Arabic}

TTS models for Arabic are typically trained with full diacritics \cite{kulkarni2023clartts}. This is because diacritics contain essential information about most vowels, without which the text is highly ambiguous. However, \citet{toyin-etal-2023-artst} demonstrated good TTS performance without the inclusion of diacritics, which is mainly attributed to the warm fine-tuning they perform with ASR data. As the TTS data include diacritics, we performed experiments where we train models with and without diacritics. We evaluate ASR  on normalized text where all diacritics are removed. 
The results are reported in Table \ref{tab:diacritics_ar}. 
We notice that performance in TTS is in fact better without diacritics using this model. This surprising observation may be attributed to the fact that ArTST\cite{toyin-etal-2023-artst} was pre-trained without diacritics, so adding diacritics in the fine-tuning stage with small data size may not be sufficient. However, as shown in the examples in Figure \ref{fig:arabcic-asr-trasncript}, we note that ASR transcription with diacritics (first line) is in fact correct, even though it does not match exactly the reference, which is mainly a result of the \textit{sukoon} diacritic that is often omitted in the reference.   

\begin{table}[h]
\centering
\ra{1.2}
\resizebox{0.8\columnwidth}{!}{%
\begin{tabular}{@{}l c cc cc@{}}
\toprule
\multirow{2}{*}{\textbf{Data}} & \multirow{2}{*}{\textbf{Diacritics}} & \multicolumn{2}{c}{\textbf{ASR}} & \multicolumn{2}{c}{\textbf{TTS}}  \\ \cmidrule(lr){3-4} \cmidrule(l){5-6} 
 & & WER $\downarrow$ & CER $\downarrow$ & CER $\downarrow$ & WV-MOS $\uparrow$\\ \midrule
$ar$ & \cmark & 10.10 & 2.75  & 7.40 & 3.68   \\
$ar$ & \xmark & 10.22  & 2.63 & 6.22 & 3.69 \\ 
 \bottomrule
\end{tabular}%
}
\caption{Results with and without diacritics.}
\label{tab:diacritics_ar}
\end{table}

\subsection{Effect of Task-Fusion Module Position}
We experimented with having the task fusion module before the encoder with the aim of guiding latent feature extraction based on the given output. This approach performs fairly well for ASR (CER 4\%) but fails for TTS with CER of 80\%.

\subsection{Effect of Pre-Trained Weights}

\begin{figure} [h]
    \centering
    \includegraphics[width=\linewidth]{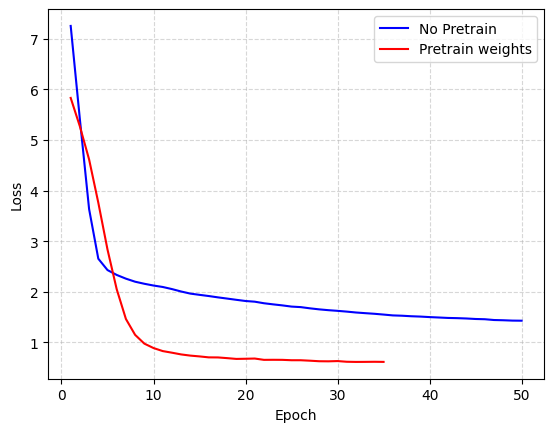}
    \caption{Comparison of training objective with and without using pre-trained weight.}
    \label{fig:pretrain-effect}
\end{figure}

The results above are all reported with pre-trained weights from SpeechT5 and ArTST pre-trained checkpoints. We examined the effect on performance with and without starting with these pre-trained weights. See Figure \ref{fig:pretrain-effect} for the learning curve in terms of loss reduction during training. We 
see that using pre-trained weights is beneficial for maximizing performance. Pre-training is particularly crucial for TTS tasks, as starting from scratch results in 10-fold degradation in CER.  On the other hand,  the performance for ASR is far less affected (only +1\% increase in CER value for \texttt{STTATTS}, and a larger increase for Y-decoder). Overall, starting with pre-trained weights is crucial in downstream tasks for all model variations. 

\section{Discussion}
We described our experiments of jointly learning Speech-to-Text and Text-to-Speech models based on the SpeechT5 architecture, resulting in a truly multi-modal and functional model, that accepts both speech and text as input and output. We experimented with both Arabic and English languages, with Arabic being a relatively low-resource language due to limited amounts of data available for TTS training. Our results show that while it is possible to train models using our framework with less than 20 hours of speech in total, more data is always better for maximizing performance in both ASR and TTS tasks. Our results are the first to report multi-task ASR/TTS in the Arabic language, showing promising results in low-resource settings, and potential for improvement with additional data. For English, a few other models have been recently proposed; our comparative analysis with the only publicly available model of this variety, named VoxtLM \cite{maiti2024VoxtLM}, favors \texttt{STTATTS} in both performance and parameter efficiency. 
We experimented with different parameter-efficient approaches to jointly learn ASR and TTS tasks and we find using the task-fusion module strikes a perfect balance in performance between both tasks with the least amount of parameters. It's also worth noting that the task-fusion module makes incorporating more tasks and output modalities feasible as the module aligns latent representation to match the desired output modality. Furthermore, the integrated multi-task approach, in addition to being more efficient in model size, is more efficient in training as it requires fewer updates in total. Future work will explore the possibility of integrating additional text output tasks within the framework and improving synthesized speech naturalness. 

We believe that the proposed model, being trained on publicly available data with the code and checkpoints publicly available\footnote{\url{https://github.com/mbzuai-nlp/sttatts}}, can serve as a strong baseline for multi-task speech processing.

\paragraph{Limitations.}

Our experiments focus on three key aspects: language generalization, scalability (in terms of training data requirements and tasks), and parameter efficiency. Although we explored two languages separately, we did not experiment with a joint multilingual models. Additionally, we used VoxLM as our only baseline for multi-task models. While other models have recently been proposed, their code and models are not yet publicly available for comparison. We did not explore increasing model size, as it would require pre-training from scratch, which is computationally expensive. However, slightly larger models could potentially enhance performance in the multi-task setting by better embedding the diverse input and output modalities. Finally, we found that subjective MOS evaluation was rather difficult to conduct as most outputs were intelligible but somewhat noisy and unnatural. For Arabic, the small data size and lack of diacritics does result in degraded intelligibility due to mismatched pronunciation of short vowels. Therefore, the reported TTS results provide some signal of quality, but may not be informative of the actual quality of the synthesized speech. 

\section*{Acknowledgements}
This work was funded in part by a Google research award, awarded in November 2023.
\bibliography{coling_latex}

\begin{thebibliography}{25}
\providecommand{\natexlab}[1]{#1}

\bibitem[{Ali et~al.(2016)Ali, Bell, Glass, Messaoui, Mubarak, Renals, and Zhang}]{ali2016mgb}
Ahmed Ali, Peter Bell, James Glass, Yacine Messaoui, Hamdy Mubarak, Steve Renals, and Yifan Zhang. 2016.
\newblock The mgb-2 challenge: Arabic multi-dialect broadcast media recognition.
\newblock In \emph{2016 IEEE Spoken Language Technology Workshop (SLT)}, pages 279--284. IEEE.

\bibitem[{Andreev et~al.(2023)Andreev, Alanov, Ivanov, and Vetrov}]{Andreev_wvmos_2023}
Pavel Andreev, Aibek Alanov, Oleg Ivanov, and Dmitry Vetrov. 2023.
\newblock \href {https://doi.org/10.1109/icassp49357.2023.10097255} {Hifi++: A unified framework for bandwidth extension and speech enhancement}.
\newblock In \emph{ICASSP 2023 - 2023 IEEE International Conference on Acoustics, Speech and Signal Processing (ICASSP)}. IEEE.

\bibitem[{Ao et~al.(2022)Ao, Wang, Zhou, Wang, Ren, Wu, Liu, Ko, Li, Zhang, Wei, Qian, Li, and Wei}]{ao2022speecht5}
Junyi Ao, Rui Wang, Long Zhou, Chengyi Wang, Shuo Ren, Yu~Wu, Shujie Liu, Tom Ko, Qing Li, Yu~Zhang, Zhihua Wei, Yao Qian, Jinyu Li, and Furu Wei. 2022.
\newblock \href {https://arxiv.org/abs/2110.07205} {Speecht5: Unified-modal encoder-decoder pre-training for spoken language processing}.
\newblock \emph{Preprint}, arXiv:2110.07205.

\bibitem[{Baevski et~al.(2020)Baevski, Zhou, Mohamed, and Auli}]{wav2vec}
Alexei Baevski, Henry Zhou, Abdelrahman Mohamed, and Michael Auli. 2020.
\newblock \href {https://arxiv.org/abs/2006.11477} {wav2vec 2.0: {A} framework for self-supervised learning of speech representations}.
\newblock \emph{CoRR}, abs/2006.11477.

\bibitem[{Bapna et~al.(2021)Bapna, an~Chung, Wu, Gulati, Jia, Clark, Johnson, Riesa, Conneau, and Zhang}]{bapna2021slam}
Ankur Bapna, Yu~an~Chung, Nan Wu, Anmol Gulati, Ye~Jia, Jonathan~H. Clark, Melvin Johnson, Jason Riesa, Alexis Conneau, and Yu~Zhang. 2021.
\newblock \href {https://arxiv.org/abs/2110.10329} {Slam: A unified encoder for speech and language modeling via speech-text joint pre-training}.
\newblock \emph{Preprint}, arXiv:2110.10329.

\bibitem[{Chung et~al.(2021)Chung, Zhang, Han, Chiu, Qin, Pang, and Wu}]{chung2021w2vbert}
Yu-An Chung, Yu~Zhang, Wei Han, Chung-Cheng Chiu, James Qin, Ruoming Pang, and Yonghui Wu. 2021.
\newblock \href {https://arxiv.org/abs/2108.06209} {W2v-bert: Combining contrastive learning and masked language modeling for self-supervised speech pre-training}.
\newblock \emph{Preprint}, arXiv:2108.06209.

\bibitem[{Devlin et~al.(2019)Devlin, Chang, Lee, and Toutanova}]{devlin2019bert}
Jacob Devlin, Ming-Wei Chang, Kenton Lee, and Kristina Toutanova. 2019.
\newblock \href {https://arxiv.org/abs/1810.04805} {Bert: Pre-training of deep bidirectional transformers for language understanding}.
\newblock \emph{Preprint}, arXiv:1810.04805.

\bibitem[{Halabi(2016)}]{halabi2016modern}
Nawar Halabi. 2016.
\newblock \emph{Modern standard Arabic phonetics for speech synthesis}.
\newblock Ph.D. thesis, University of Southampton.

\bibitem[{Houlsby et~al.(2019)Houlsby, Giurgiu, Jastrzebski, Morrone, De~Laroussilhe, Gesmundo, Attariyan, and Gelly}]{pmlr-v97-houlsby19adapter}
Neil Houlsby, Andrei Giurgiu, Stanislaw Jastrzebski, Bruna Morrone, Quentin De~Laroussilhe, Andrea Gesmundo, Mona Attariyan, and Sylvain Gelly. 2019.
\newblock \href {https://proceedings.mlr.press/v97/houlsby19a.html} {Parameter-efficient transfer learning for {NLP}}.
\newblock In \emph{Proceedings of the 36th International Conference on Machine Learning}, volume~97 of \emph{Proceedings of Machine Learning Research}, pages 2790--2799. PMLR.

\bibitem[{Hsu et~al.(2021)Hsu, Bolte, Tsai, Lakhotia, Salakhutdinov, and Mohamed}]{hsu2021hubert}
Wei-Ning Hsu, Benjamin Bolte, Yao-Hung~Hubert Tsai, Kushal Lakhotia, Ruslan Salakhutdinov, and Abdelrahman Mohamed. 2021.
\newblock Hubert: Self-supervised speech representation learning by masked prediction of hidden units.
\newblock \emph{IEEE/ACM Transactions on Audio, Speech, and Language Processing}, 29:3451--3460.

\bibitem[{Ito and Johnson(2017)}]{ljspeech17}
Keith Ito and Linda Johnson. 2017.
\newblock The lj speech dataset.
\newblock \url{https://keithito.com/LJ-Speech-Dataset/}.

\bibitem[{Kominek and Black(2004)}]{kominek04b_cmuarctic}
John Kominek and Alan~W. Black. 2004.
\newblock The cmu arctic speech databases.
\newblock In \emph{5th ISCA Workshop on Speech Synthesis (SSW 5)}, pages 223--224.

\bibitem[{Kulkarni et~al.(2023)Kulkarni, Kulkarni, Shatnawi, and Aldarmaki}]{kulkarni2023clartts}
Ajinkya Kulkarni, Atharva Kulkarni, Sara Abedalmonem~Mohammad Shatnawi, and Hanan Aldarmaki. 2023.
\newblock \href {https://arxiv.org/abs/2303.00069} {Clartts: An open-source classical arabic text-to-speech corpus}.
\newblock \emph{Preprint}, arXiv:2303.00069.

\bibitem[{Maiti et~al.(2024)Maiti, Peng, Choi, weon Jung, Chang, and Watanabe}]{maiti2024VoxtLM}
Soumi Maiti, Yifan Peng, Shukjae Choi, Jee weon Jung, Xuankai Chang, and Shinji Watanabe. 2024.
\newblock \href {https://arxiv.org/abs/2309.07937} {Voxtlm: unified decoder-only models for consolidating speech recognition/synthesis and speech/text continuation tasks}.
\newblock \emph{Preprint}, arXiv:2309.07937.

\bibitem[{Panayotov et~al.(2015)Panayotov, Chen, Povey, and Khudanpur}]{Librispeech}
Vassil Panayotov, Guoguo Chen, Daniel Povey, and Sanjeev Khudanpur. 2015.
\newblock \href {https://doi.org/10.1109/ICASSP.2015.7178964} {Librispeech: An asr corpus based on public domain audio books}.
\newblock In \emph{2015 IEEE International Conference on Acoustics, Speech and Signal Processing (ICASSP)}, pages 5206--5210.

\bibitem[{Radford et~al.(2022)Radford, Kim, Xu, Brockman, McLeavey, and Sutskever}]{whisperradford2022robust}
Alec Radford, Jong~Wook Kim, Tao Xu, Greg Brockman, Christine McLeavey, and Ilya Sutskever. 2022.
\newblock \href {https://arxiv.org/abs/2212.04356} {Robust speech recognition via large-scale weak supervision}.
\newblock \emph{Preprint}, arXiv:2212.04356.

\bibitem[{Snyder et~al.(2018)Snyder, Garcia-Romero, Sell, Povey, and Khudanpur}]{x-vectors}
David Snyder, Daniel Garcia-Romero, Gregory Sell, Daniel Povey, and Sanjeev Khudanpur. 2018.
\newblock \href {https://doi.org/10.1109/ICASSP.2018.8461375} {X-vectors: Robust dnn embeddings for speaker recognition}.
\newblock In \emph{2018 IEEE International Conference on Acoustics, Speech and Signal Processing (ICASSP)}, pages 5329--5333.

\bibitem[{Tachibana et~al.(2018)Tachibana, Uenoyama, and Aihara}]{8461829}
Hideyuki Tachibana, Katsuya Uenoyama, and Shunsuke Aihara. 2018.
\newblock \href {https://doi.org/10.1109/ICASSP.2018.8461829} {Efficiently trainable text-to-speech system based on deep convolutional networks with guided attention}.
\newblock In \emph{2018 IEEE International Conference on Acoustics, Speech and Signal Processing (ICASSP)}, pages 4784--4788.

\bibitem[{Toyin et~al.(2023)Toyin, Djanibekov, Kulkarni, and Aldarmaki}]{toyin-etal-2023-artst}
Hawau Toyin, Amirbek Djanibekov, Ajinkya Kulkarni, and Hanan Aldarmaki. 2023.
\newblock \href {https://aclanthology.org/2023.arabicnlp-1.5} {{A}r{TST}: {A}rabic text and speech transformer}.
\newblock In \emph{Proceedings of ArabicNLP 2023}, pages 41--51, Singapore (Hybrid). Association for Computational Linguistics.

\bibitem[{Vaswani et~al.(2017)Vaswani, Shazeer, Parmar, Uszkoreit, Jones, Gomez, Kaiser, and Polosukhin}]{NIPS2017_attention}
Ashish Vaswani, Noam Shazeer, Niki Parmar, Jakob Uszkoreit, Llion Jones, Aidan~N Gomez, \L~ukasz Kaiser, and Illia Polosukhin. 2017.
\newblock \href {https://proceedings.neurips.cc/paper_files/paper/2017/file/3f5ee243547dee91fbd053c1c4a845aa-Paper.pdf} {Attention is all you need}.
\newblock In \emph{Advances in Neural Information Processing Systems}, volume~30. Curran Associates, Inc.

\bibitem[{Wang et~al.(2023)Wang, Zhou, Zhang, Wu, Liu, Gaur, Chen, Li, and Wei}]{wang2023viola}
Tianrui Wang, Long Zhou, Ziqiang Zhang, Yu~Wu, Shujie Liu, Yashesh Gaur, Zhuo Chen, Jinyu Li, and Furu Wei. 2023.
\newblock \href {https://arxiv.org/abs/2305.16107} {Viola: Unified codec language models for speech recognition, synthesis, and translation}.
\newblock \emph{Preprint}, arXiv:2305.16107.

\bibitem[{Watanabe et~al.(2018)Watanabe, Hori, Karita, Hayashi, Nishitoba, Unno, Soplin, Heymann, Wiesner, Chen, Renduchintala, and Ochiai}]{espnet}
Shinji Watanabe, Takaaki Hori, Shigeki Karita, Tomoki Hayashi, Jiro Nishitoba, Yuya Unno, Nelson Enrique~Yalta Soplin, Jahn Heymann, Matthew Wiesner, Nanxin Chen, Adithya Renduchintala, and Tsubasa Ochiai. 2018.
\newblock \href {https://arxiv.org/abs/1804.00015} {Espnet: End-to-end speech processing toolkit}.
\newblock \emph{CoRR}, abs/1804.00015.

\bibitem[{Zen et~al.(2019)Zen, Dang, Clark, Zhang, Weiss, Jia, Chen, and Wu}]{zen2019libritts}
Heiga Zen, Viet Dang, Rob Clark, Yu~Zhang, Ron~J. Weiss, Ye~Jia, Zhifeng Chen, and Yonghui Wu. 2019.
\newblock \href {https://arxiv.org/abs/1904.02882} {Libritts: A corpus derived from librispeech for text-to-speech}.
\newblock \emph{Preprint}, arXiv:1904.02882.

\bibitem[{Zhang et~al.(2023{\natexlab{a}})Zhang, Li, Zhang, Zhan, Wang, Zhou, and Qiu}]{zhang2023speechgpt}
Dong Zhang, Shimin Li, Xin Zhang, Jun Zhan, Pengyu Wang, Yaqian Zhou, and Xipeng Qiu. 2023{\natexlab{a}}.
\newblock \href {https://arxiv.org/abs/2305.11000} {Speechgpt: Empowering large language models with intrinsic cross-modal conversational abilities}.
\newblock \emph{Preprint}, arXiv:2305.11000.

\bibitem[{Zhang et~al.(2023{\natexlab{b}})Zhang, Han, Qin, Wang, Bapna, Chen, Chen, Li, Axelrod, Wang, Meng, Hu, Rosenberg, Prabhavalkar, Park, Haghani, Riesa, Perng, Soltau, Strohman, Ramabhadran, Sainath, Moreno, Chiu, Schalkwyk, Beaufays, and Wu}]{zhang2023googleusm}
Yu~Zhang, Wei Han, James Qin, Yongqiang Wang, Ankur Bapna, Zhehuai Chen, Nanxin Chen, Bo~Li, Vera Axelrod, Gary Wang, Zhong Meng, Ke~Hu, Andrew Rosenberg, Rohit Prabhavalkar, Daniel~S. Park, Parisa Haghani, Jason Riesa, Ginger Perng, Hagen Soltau, Trevor Strohman, Bhuvana Ramabhadran, Tara Sainath, Pedro Moreno, Chung-Cheng Chiu, Johan Schalkwyk, Françoise Beaufays, and Yonghui Wu. 2023{\natexlab{b}}.
\newblock \href {https://arxiv.org/abs/2303.01037} {Google usm: Scaling automatic speech recognition beyond 100 languages}.
\newblock \emph{Preprint}, arXiv:2303.01037.

\end{thebibliography}

\end{document}